\documentclass[titlepage,twocolumn,a4paper]{article}
\usepackage[pdftex]{graphicx,color}
\usepackage{enumitem}
\usepackage{multirow}
\usepackage{amssymb, amsmath,mathtools}
\usepackage{latexsym}
\usepackage{bm}
\usepackage{breqn}
\usepackage{color}
\usepackage{ulem}
\usepackage{ascmac}
\usepackage{comment}
\usepackage{siunitx}
\usepackage{framed}
\usepackage{here}
\usepackage{float}
\usepackage{array}
\usepackage{indentfirst}
\newcolumntype{C}{>{\centering\arraybackslash}m{3cm}}
\usepackage{titlesec}
\titlespacing*{\section}{0pt}{\baselineskip}{\baselineskip}
\titlespacing*{\subsection}{0pt}{\baselineskip}{\baselineskip}
\newcommand{\x}{{\bm{x}}}
\newcommand{\y}{{\bm{y}}}
\newcommand{\z}{{\bm{z}}}
\newcommand{\bR}{{\mathbb{R}}}
\usepackage[hang,small,bf]{caption}
\usepackage[subrefformat=parens]{subcaption}

\begin{document}

\twocolumn[%
\begin{center}
\text{\Large{A Study on Unsupervised Anomaly Detection and Defect Localization}} \\ \text{\Large{using Generative Model in Ultrasonic Non-Destructive Testing}}

\vspace{0.35cm}

\large
\mbox{}
\begin{tabular}{cccc}
  {\sffamily Yusaku Ando$^a$}  & {\sffamily Miya Nakajima$^a$}  & {\sffamily Takahiro Saitoh$^a$} & {\sffamily Tsuyoshi Kato$^b$} 
\end{tabular}
\mbox{}

\vspace{0.35cm}

\small
${}^a$\textit{Graduate School of Science and Technology, Gunmma University, Tenjin, Kiryu 376 8515, Japan,}\\
${}^b$\textit{Faculty of Informatics, Gunmma University, Aramaki, Maebashi 371 8510, Japan.}

\end{center}

\vspace{0.5cm}

\hrulefill

\vspace{0.2cm}
\textbf{Abstract} \vspace{0.2cm} \\
\noindent In recent years, the deterioration of artificial materials used in structures has become a serious social issue, increasing the importance of inspections.
Non-destructive testing is gaining increased demand due to its capability to inspect for defects and deterioration in structures while preserving their functionality.
Among these, Laser Ultrasonic Visualization Testing (LUVT) stands out because it allows the visualization of ultrasonic propagation. 
This makes it visually straightforward to detect defects, thereby enhancing inspection efficiency.
With the increasing number of the deterioration structures, challenges such as a shortage of inspectors and increased workload in non-destructive testing have become more apparent.
Efforts to address these challenges include exploring automated inspection using machine learning.
However, the lack of anomalous data with defects poses a barrier to improving the accuracy of automated inspection through machine learning.
Therefore, in this study, we propose a method for automated LUVT inspection using an anomaly detection approach with a diffusion model that can be trained solely on negative examples (defect-free data).
We experimentally confirmed that our proposed method improves defect detection and localization compared to general object detection algorithms used previously.

\vspace{0.2cm}


\hrulefill

\medskip]

\section{Introduction}

\indent
In recent years, the deterioration of artificial materials that constitute social infrastructure such as dams and bridges has become a serious societal issue, highlighting the increasing importance of maintenance and management.
Non-destructive testing, which allows the inspection of defects or deterioration without destroying the object under examination, is gaining increased demand due to its advantages in inspection efficiency and safety.
In non-destructive testing, internal defects and their locations are examined using techniques such as radiation or ultrasonic waves.
Among them, laser ultrasonic visualization testing (LUVT)~\cite{auto-ndt-surp-saitoh,auto-ndt-surp-nakajima,luvt-da-nakajima}, which uses laser ultrasonic to easily inspect objects with complex shapes, is attracting attention.
The research and development of LUVT technology are advancing because it allows the visualization of the propagation of ultrasonic, making it visually clear to determine the presence of defects~\cite{XIE2024110876}.
However, against the backdrop of a deteriorating structures in recent years, issues such as shortages of inspectors and increased workloads have arisen.

On the other hand, there has been remarkable development in the field of deep learning in recent years, enabling models to achieve performance in visual tasks such as object detection and recognition that matches or even exceeds human capabilities.
Therefore, there is ongoing consideration of utilizing machine learning techniques to automate inspections for assessing anormalities in non-destructive testing~\cite{auto-ndt-surp-saitoh,auto-ndt-pinn,MENG2017128}.
However, a problem in applying machine learning techniques to automate non-destructive testing is the difficulty in obtaining sufficient positive (defective) data for training.
Furthermore, even if there is a large quantity of positive training data available, the labeling process for these samples demands significant time and effort, which is also problematic.
To address these problems, data augmentation techniques~\cite{luvt-da-nakajima} and anomaly detection methods~\cite{HA2022106637} have been proposed.
However, as far as the authors know, there are no precedents of applying unsupervised learning in LUVT research, with studies predominantly focusing on supervised learning approaches.

Therefore, in this paper, we propose a method for automated LUVT inspection using AnoDDPM~\cite{Wyatt_2022_CVPR}, an anomaly detection approach based on diffusion model.
In our proposed method, we utilize time-series information consisting only of normal data to conduct inspections.
This approach not only enables defect detection but also allows for the estimation of defect locations.
In this paper, we indicate this methodology and validate the effectiveness of our proposed approach by comparing it with several supervised learning methods.

\section{Related Work}
\subsection{History of Machine Learning Adoption in Ultrasonic Non-Destructive Testing}

The application of machine learning in ultrasonic non-destructive testing has been increasing due to its capability to automatically inspect large volumes of data.
In the early stages, simple algorithms such as Support Vector Machine (SVM) or Fisher Linear Discriminant were used for defect detection and classification in ultrasonic non-destructive testing~\cite{MASNATA199687,baker1989classification,damarla1992self}.
In recent years, with developments in deep learning algorithms and improvements in computational power, more complex machine learning models such as Convolutional Neural Networks (CNN) and Deep Neural Networks (DNN) have emerged.
These models have enabled the performance of various tasks in ultrasonic non-destructive testing with high accuracy.
In image classification tasks, Virupakshappa \textit{et al}. performed defect detection on A-scan data obtained from steel materials using SVM-based image classification~\cite{surp-svm}.
Additionally, Cruz \textit{et al}. applied techniques such as discrete Fourier transform to A-scan data for detecting welding defects, followed by image classification using a multilayer perceptron~\cite{CRUZ20171}.

Furthermore, research has been conducted to apply object detection and segmentation tasks not only for detecting the presence of defects but also for identifying defect locations.
In object detection tasks, Gantala \textit{et al}. applied YOLOv4, a model designed for object detection, to TFM images obtained from welded carbon steel to perform defect detection~\cite{gantala2021automated}.
Additionally, Medak \textit{et al}. improved defect detection accuracy and reduced inference time by improving components of EfficientDet, which enables efficient object detection~\cite{MEDAK2022107}.
Moreover, Provencal \textit{et al}. applied segmentation tasks to S-scan data obtained from carbon steel to identify welding shapes~\cite{PROVENCAL2021122}.
Caballero \textit{et al}. automated the segmentation of ultrasonic data on Carbon Fiber Reinforced Plastics (CFRP) using X-ray Computed Tomography (XCT) data and CNN~\cite{app13105933}.
However, these studies are based on supervised learning, and applying them directly to non-destructive testing raises the following issues.
\begin{enumerate}
  \item Collecting the necessary anormal data for training is challenging.
  \item Paired data consisting of images and corresponding (per region) labels are required as training data.
  \item Unable to make correct decisions when encountering unknown anomalies.
\end{enumerate}

Therefore, unsupervised learning has been introduced to the field of non-destructive inspection as it does not depend on teacher signals.
Unsupervised learning-based automated inspection is broadly categorized into defect detection using generative model or feature extraction.
For example, in approaches utilizing generative model, Ha \textit{et al}. attempted defect detection near the surface using autoencoder~\cite{HA2022106637}.
Posilovi\'{c} \textit{et al}. improved the Ganomaly model, which is an anomaly detection method using GAN, to apply it to ultrasonic non-destructive testing~\cite{POSILOVIC2022106737}.
Momreover, example in feature extraction-based approaches is the study by Bowler \textit{et al}., who applied principal component analysis to features extracted from ultrasonic waveforms using CNN~\cite{BOWLER2021107508}.

The LUVT focused on in this research is still a new technology, and attempts to apply machine learning have not been fully explored.
Examples of previous research include a study by Saitoh \textit{et al}. using CNN for automatic inspection of concrete materials~\cite{auto-ndt-surp-saitoh}, and another study by Nakajima \textit{et al}. where they improved the CNN architecture for automated LUVT inspection~\cite{auto-ndt-surp-nakajima}.
These are examples of supervised learning applications and do not resolve the issues of supervised learning in ultrasonic non-destructive testing mentioned earlier.

\subsection{Anomaly Detection}

An example of a learning method that does not require anormal data is unsupervised anomaly detection.
Unsupervised anomaly detection methods are primarily classified into three categories: synthetic-based methods, embedding-based methods, and generative model-based methods.

In synthetic-based methods, anomalies are generated by synthesizing them into normal images to create pseudo-anomalous images.
For example, in CutPaste~\cite{CutPaste}, pseudo-anomalous images are generated by cutting out a portion of a normal image and pasting it onto another location within the same image.
During training, CNN are trained using available normal images along with the generated pseudo-anomalous images.
However, it has been noted that synthetic-based methods may not cover all anomalies comprehensively, and the generated pseudo-anomalous images may sometimes deviate significantly from actual anomalies~\cite{he2023diad,Liu_2023_CVPR}.

In embedding-based methods, pre-trained CNN, often trained on ImageNet, are used to extract features from normal images.
PaDiM~\cite{PaDiM} assumes a multivariate normal distribution for feature extraction, while PatchCore~\cite{PatchCore} stores normal image features in a memory bank.
During testing, anomaly detection is determined by using Mahalanobis distance or maximum feature distance to assess whether the input image is anormal or not.
However, using features pre-trained on datasets like ImageNet, which consist of natural images, for industrial anomaly detection tasks may lead to difficulties in accurate detection.
Additionally, there are concerns about high memory consumption associated with these methods~\cite{he2023diad,Liu_2023_CVPR}.

In generative model-based methods, the approach is based on the idea that using a generative model trained only on normal data can reconstruct anomalous image regions as if they were normal.
The generative models commonly known for achieving this include VAE (Variational Autoencoder)~\cite{kingma2022autoencoding}, GAN (Generative Adversarial Networks)~\cite{goodfellow2014generative}, and Diffusion model~\cite{DDPM}.
In generative model-based methods, anomaly detection is performed by creating an anomaly map through the difference between the input image and the reconstructed image, enabling defect detection.
In recent years, generative models have begun to be applied in various fields due to their ability to generate high-quality images.
In this study, considering the promising potential of generative model-based methods, we have adopted a generative model-based anomaly detection approach.

\section{LUVT Dataset}
\subsection{What is LUVT?}

\begin{figure}[hbtp]
  \centering
  \includegraphics[keepaspectratio, scale=1.25]{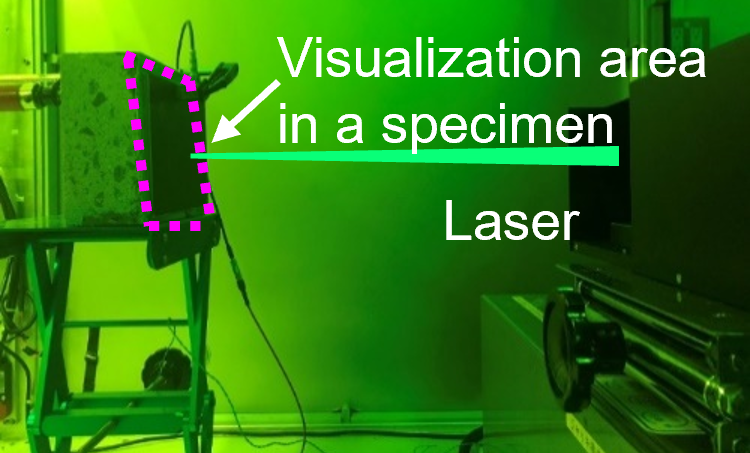}
  \caption{LUVT device.}
  \label{fig:luvt-eq}
 \end{figure}

\begin{figure}[hbtp]
   \begin{tabular}{cc}
     \rm (a)Image without defect & \rm (b)Image with defect \\
     \begin{minipage}[t]{0.45\hsize}
       \centering
       \includegraphics[keepaspectratio, scale=0.65]{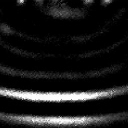}
     \end{minipage} &
     \begin{minipage}[t]{0.45\hsize}
       \centering
       \includegraphics[keepaspectratio, scale=0.65]{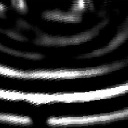}
     \end{minipage} \\
   \end{tabular}
  \caption{Examples of LUVT images. (a) is an example of an image without defect, while (b) is an example of an image with a defect at the bottom. In the image without defects, it is evident that waves propagate from the top of the image. In the image with a defect, anormalities in the waves occur because the incident wave encounters the defect.}
  \label{fig:ex-luvt}
\end{figure}

In non-destructive testing using LUVT, as shown in Figure \ref{fig:luvt-eq}, a pulsed laser is directed onto the surface of the material under examination, where it generates and propagates ultrasonic waves at the laser irradiation point (transmit point).
At this point, ultrasonic waves are received by an ultrasonic transducer previously installed on the test specimen.
Using the reciprocal theorem, the temporal history image of the ultrasonic wave propagation from the ultrasonic transducer can be visualized.
This procedure is performed on an aluminum test specimen, and an example image obtained is shown in Figure \ref{fig:ex-luvt}.
In Figure \ref{fig:ex-luvt}(a), which represents an image with no defects, the incident wave from the transducer propagates without interruption.
On the other hand, in the case shown in Figure \ref{fig:ex-luvt}(b) where a defect exists, when the incident wave reaches the defect, an anormality occurs in the wave.
By identifying such changes in the wave, it becomes possible for individuals, even those without extensive expertise, to detect defects in the material.

\subsection{Creating Dataset}

The test specimen used in LUVT is an aluminum specimen measuring 30 mm in vertical, 100 mm in horizontal, and 30 mm in depth.
The LUVT images shown in Figure \ref{fig:ex-luvt} was obtained by irradiating a laser onto an arbitrary rectangular area measuring 20 mm in vertical and 50 mm in horizontal within the test specimen.
In this study, we propose a method that does not use defective data.
However, defective data is required for performance evaluation and learning of conventional approaches.
Therefore, defective data was obtained by artificially creating cylindrical voids with a diameter of 2 mm in the aluminum test specimen.
In LUVT, for each test, a sequence of temporal history images (video) depicting the propagation of ultrasonic at the laser irradiation area is obtained.
The breakdown of data includes 55 videos of defective data and 123 videos of defect-free data.
Details of the dataset used in the experiments will be described in Section \ref{exp}.
The transducer used was a longitudinal wave transducer with a center frequency of \SI{2}{MHz}.

\section{Proposal Method}
\subsection{Generative Models}

In this section, we will briefly introduce the generative models (VAE, GAN, Diffusion model) that were considered for use in the proposed method.
The proposed method performs anomaly detection using generative models trained only on normal images.
The inference of defect detection leverages the property of generative models, where the defective parts are reconstructed.

\subsubsection{VAE (Variational Autoencoder)}

VAE consists of an encoder and a decoder.
In the encoder, instead of directly finding the latent variable $\z$ for the input data $\x$, it is assumed that the latent variable is sampled from a certain probability distribution. 
The encoder then estimates the mean and variance of this distribution.
In the decoder, given a sampled latent variable $\z$, the decoder outputs a reconstructed data $\x^{\prime}$ that restores the input data.
The encoder is represented by an approximate posterior distribution $q_{\boldsymbol{\phi}}(\z|\x)$, and the decoder is represented by a probabilistic generative model $p_{\boldsymbol{\theta}}(\x|\z)$. 
In VAE, we explore  to maximize the following variational lower bound by searching for learnable parameters $\boldsymbol{\phi}$ and $\boldsymbol{\theta}$:
\begin{align}
  &\mathcal{L}_{\text{VAE}} \notag \\
  &= -\mathbb{E}_{q_{\boldsymbol{\phi}}(\z|\x)} \left[ \log p_{\boldsymbol{\theta}}(\x|\z) \right] + D_{\text{KL}}\left[ q_{\boldsymbol{\phi}}(\z|\x) \| p(\z ) \right]. \label{eq:vae_loss}
\end{align}
Furthermore, to enable analytical optimization of equation (\ref{eq:vae_loss}), the prior distribution $p(\z)$ and the approximate posterior distribution $q_{\boldsymbol{\phi}}(\z|\x)$ are modeled as normal distribution. 
Typically, the prior distribution is assumed to be a standard normal distribution $\mathcal{N}(0,1)$.

\subsubsection{GAN (Generative Adversarial Networks)}

GAN typically consists of two networks: the Generator and the Discriminator.
The Generator, when given a latent variable $\z$, generates data using the learned data distribution $p_g$.
The distribution $p_{g}$ of the Generator is defined by a prior distribution $p_{\z}(\z)$ of random noise $\z$ and a mapping $G(\z,\boldsymbol{\theta}_{g})$ from the latent space to the data space, and it does training.
The Discriminator, when given a certain data, outputs the probability that the data was generated not from the real data distribution (true data distribution) but from the distribution $G(\z,\boldsymbol{\theta}_{g})$ of the Generator.
In practice, it defines $D(\x,\boldsymbol{\theta}_{d})$ and outputs 0 or 1.
In other words, the Discriminator identifies whether the input data is generated data or real data.
The Generator and Discriminator each have their own objectives: the Generator aims to deceive the Discriminator by generating data that appears real, while the Discriminator aims to correctly distinguish between generated data and real data.
The loss function during training of a GAN is given by the following equation:
\begin{multline}
\underset{G}{\min}\ \underset{D}{\max}\ V(D,G) = \mathbb{E}_{\x \sim p_{\text{data}(\x)}}\left[ \log D(\x) \right] \\ 
+ \mathbb{E}_{\z\sim{p_{\z}(\z)}}\left[ \log(1 - D(G(\z))) \right]. \label{eq:gan-loss}
\end{multline}

\subsubsection{Diffusion Model}

The diffusion model consists of a Forward Process and a Reverse Process.
Forward Process is the process of adding noise to the input image.
Reverse Process is the process of reconstructing the input image from noise.

The Forward Process involves repeatedly adding Gaussian noise to the input image $\x_{0}\in\bR^{h\times w}$, where $w$ and $h$ represent the width and height of the image, respectively.
Let us denote each image at each step as $\x_{0},\dots,\x_{T}$.
The distribution of $\x_{t}$ generated from $\x_{t-1}$ can be expressed as follows:
\begin{align}
  q(\x_{t}\,|\,\x_{t-1}) := 
  \mathcal{N}(\x_{t}\,;\,\sqrt{1-\beta_{t}}\x_{t-1},\beta_{t}\bm{I}). 
\end{align}
In this case, $\beta_{t}$ represents a constant indicating the magnitude of the noise. 
Therefore, according to the properties of the normal distribution,
\begin{align}
  \x_{t}=\sqrt{\bar{\alpha}_{t}}\x_{0} + \sqrt{1-\bar{\alpha}_{t}}\boldsymbol{\epsilon}, 
  \quad \boldsymbol{\epsilon}\sim\mathcal{N}(\bm{0},\bm{I}),
\end{align}
where $\bar{\alpha}_{t}:=\prod_{s=1}^{t}(1-\beta_{s})$.

In the Reverse Process, we attempt to gradually restore the input images $\x_{T},\x_{T-1},\dots,\x_{0}$ from the noise $\x_{T}$ using the following distribution:
\begin{align}
  p_{\boldsymbol{\theta}}(\x_{t-1}\,|\,\x_{t})=\mathcal{N}\left( \x_{t-1}\,;\,\boldsymbol{\mu}_{\boldsymbol{\theta}}(\x_{t},t),\frac{1-\bar{\alpha}_{t-1}}{1-\bar{\alpha}_{t}}{\beta}_{t}\bm{I}\right).
\end{align}
The vector-valued function $\boldsymbol{\mu}_{\boldsymbol{\theta}}$ is constructed using a neural network called U-Net, and its weight parameters $\boldsymbol{\theta}$ are trained to minimize the KL divergence between distributions $q$ and $p$ as follows:
\begin{multline}
  L = \mathbb{E}_{q}\bigg[ D_{\text{KL}}(q(\x_{T}|\x_0)  \| p(\x_T)) \bigg. \\ 
  \bigg. + \sum_{t>1} D_{\text{KL}}(q(\x_{t-1}|\x_t,\x_0)) - \log p_{\boldsymbol{\theta}}(\x_0|\x_1) \bigg]. \label{eq:loss-ddpm}
\end{multline}

\subsection{Defect Localization Method}

\begin{figure}[hbtp]
 \centering
 \includegraphics[keepaspectratio, scale=0.35]{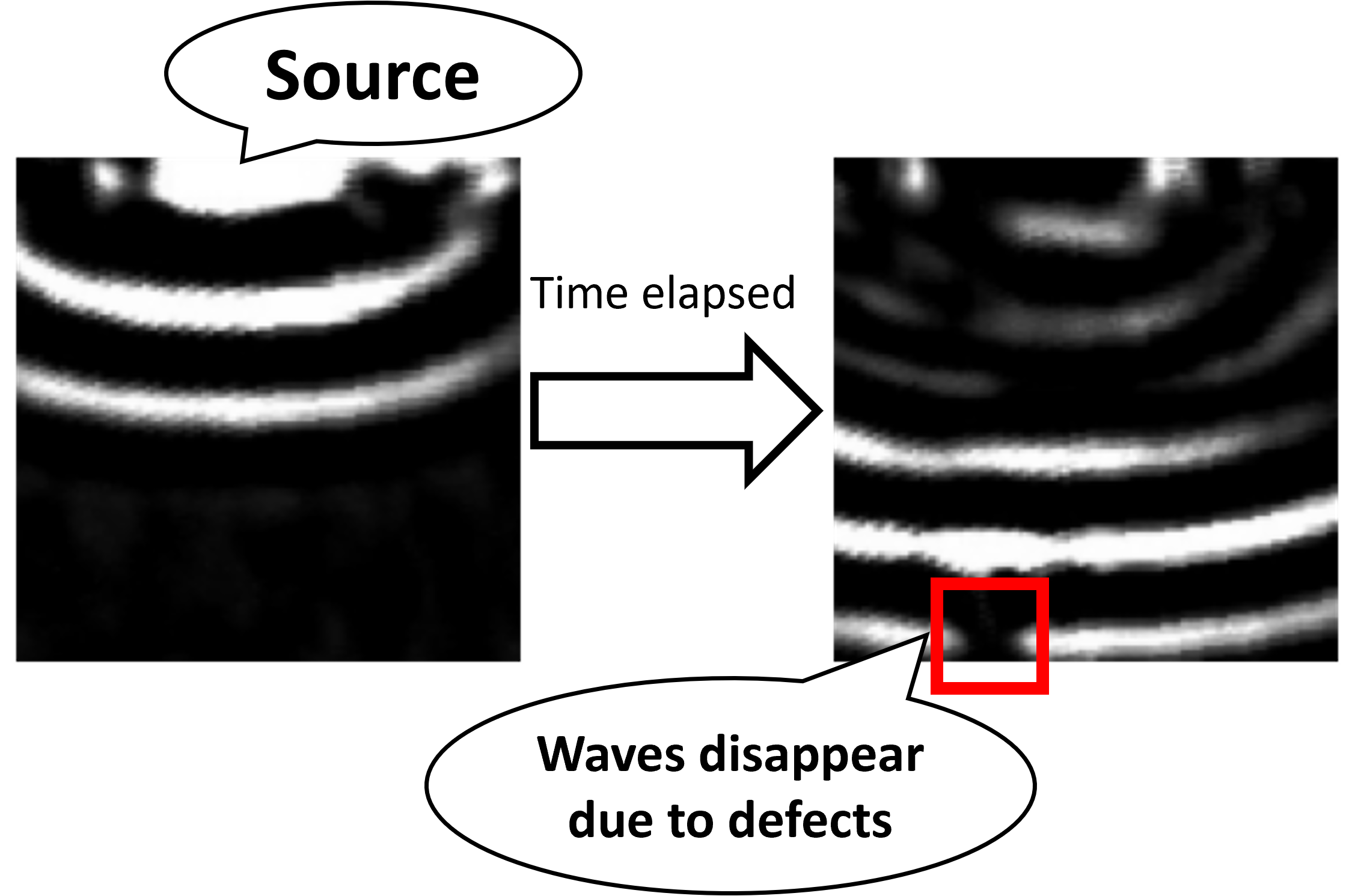}
 \caption{Appearance of ultrasonic before and after reaching a defect.}
 \label{fig:loss-wave}
\end{figure}

LUVT is a technique that uses signals obtained by irradiating the test object with ultrasonic to construct a sequence of images depicting the propagation of ultrasonic within the object.
When there is a defect inside the test object, anormalities occur in the propagation of ultrasonic (as shown in Figure \ref{fig:loss-wave}).
The location of the defect can be identified from the anormalities in the ultrasonic propagation. 

\begin{figure}[hbtp]
  \centering
  \includegraphics[keepaspectratio, scale=0.25]{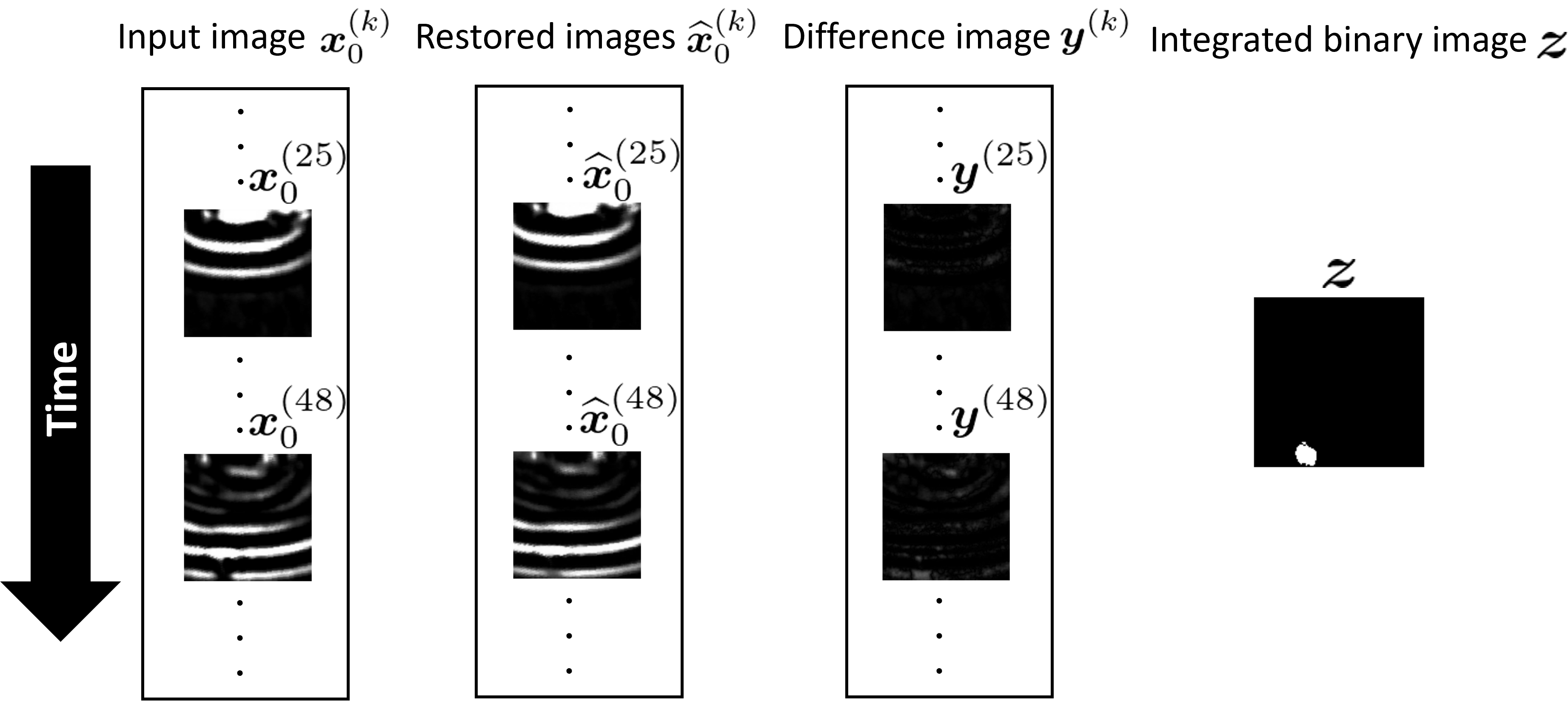}
  \caption{Flow of the proposed method.}
  \label{fig:proposal-flow}
\end{figure}
In this study, we constructed an algorithm to estimate the defect location from the LUVT image sequence as follows (as shown in Figure \ref{fig:proposal-flow}).
Givng a sequence of $K$ frames of images $\x_{0}^{(1)},\dots,\x_{0}^{(K)}\in\bR^{h\times w}$.
Let $\widehat{\x}_{0}^{(k)}\in\bR^{h\times w}$ be the image restored by applying the diffusion model to each image $\x_{0}^{(k)}\in\bR^{h\times w}$.
Compute the difference image $\y^{(k)}:=|\x_{0}^{(k)}-\widehat{\x}_{0}^{(k)}|\in\bR^{h\times w}$.
Then, you obtain the difference images $\y^{(1)},\dots,\y^{(K)}$ for the $K$ frames.
From the $K$ frames of images, you obtain a single binary image $\z\in\{0,1\}^{h\times w}$ that integrates the highlighted defects.
In the binary image $\z\in\{0,1\}^{h\times w}$, a luminance value of $1$ represents the presence of a defect, while $0$ indicates the absence of a defect.
For the connected components in the binary image $\z$, we predict a defect for those components where the luminance value before binarization exceeds a threshold (set to 1000 in this experiment).
Also, when the sum of the luminance values before binarization for the largest connected component is less than the threshold, we classify it as defect-free.
In the experiments described later, we evaluated performance by using the centroid of the largest area connected component as the estimated defect position.

In this study, the binary image $\z$ was obtained through a two-stage binarization process as follows:
In the first stage, binary images were computed for each of the $K$ difference images $\y^{(1)},\dots,\y^{(K)}$ using multiple thresholds.
The mean image $\z_{\text{pre}}\in\bR^{h\times w}$ was obtained by calculating the average of these binary images, followed by another binarization process to obtain $\z$.
For the first-stage binarization, we used 21 thresholds ranging from $0.15, 0.16, \dots, 0.35$.
In this case, we would obtain $21K$ binary images.
For the second-stage binarization, we used a threshold of one-third of the maximum luminance value among the $w \cdot h$ pixel values in the average image $\z_{\text{pre}}$ obtained from the $21K$ binary images.

\section{Exploration of Generative Models and Performance Evaluation for Defect Detection}\label{exp}

In this study, we first conducted preliminary experiments to determine which generative model is most suitable for implementing the proposed method for automated LUVT inspection.
Next, we conducted evaluation experiments to compare the proposed method's defect detection and localization performance with that of conventional approaches, specifically general object detection algorithms conventionally used for automated LUVT inspection.
In LUVT, aluminum was used as the inspection material, and the image size during training was set to 128$\times$128 pixels in grayscale.

\subsection{Overview of the Comparative Experiments on Generative Models}\label{com-gen-models-pre}

We compared VAE~\cite{kingma2022autoencoding}, f-AnoGAN~\cite{SCHLEGL201930}, and AnoDDPM~\cite{Wyatt_2022_CVPR} as generative models.
In this section, we applied the generative model trained solely on normal data to LUVT images.
By using the difference between the input and output of the generative model as the anomaly score, we evaluated the performance using AUROC (Area Under Receiver Operating Characteristic).
The dataset used for training and testing was structured as shown in Table \ref{tab:gen-model-data}.
\begin{table}[hbtp]
  \caption{Dataset configuration for the generative model comparison experiment.}
  \label{tab:gen-model-data}
  \centering
  \begin{tabular}{c|cc}
    \hline
      & Defective & Defect-free \\ \hline
      Training set & 6024 & 0     \\
      Testing set & 500 & 500    \\ \hline
  \end{tabular}
\end{table}

The experimental settings for each generative model were set as follows.

\subsubsection{VAE}

For VAE~\cite{kingma2022autoencoding}, we used 200 epochs with the Adam optimiser and a learning rate of $1.0\times10^{-3}$.
The loss function used was mean squared error.

\subsubsection{f-AnoGAN}

For f-AnoGAN~\cite{SCHLEGL201930}, we used 300 epochs for GAN training and 200 epochs for the encoder training. 
Both used the Adam optimiser with a learning rate of $2.0\times10^{-4}$.
Additionally, we employed the izif architecture, which was reported as the most effective in literature reference~\cite{SCHLEGL201930}.

\subsubsection{AnoDDPM}\label{com-gen-models-pre-ddpm}

The model structure of AnoDDPM used in this study follows the approach outlined in literature reference ~\cite{Wyatt_2022_CVPR}.
The noise used in this study was the simplex noise employed in literature reference~\cite{Wyatt_2022_CVPR}.
Also, the number of training steps was set to 1000, and the noise scheduler used the cosine schedule that yielded good results in literature reference~\cite{improvedddpm}.
In addition, the number of epochs was set to 1000, the optimiser used was AdamW, and the learning rate was set to $1.0 \times 10^{-6}$.

\subsection{Evaluation Metrics for the Comparison Experiment on Generative Models}

In this section, we use AUROC for evaluation.
AUROC refers to the area under of the curve at ROC curve which plots the true positive rate (TPR) on the vertical axis against the false positive rate (FPR) on the horizontal axis.
If the classification is entirely correct, the AUROC will be 1.0.
Also, the AUROC of 0.5 means that the performance is equivalent to random guessing.

The anomaly score was calculated by binarizing the absolute error between the input image $\x$ and the reconstruction image $f(\x)$.
We evaluate the performance by varying the threshold for the anomaly score used to determine anomalies.
Threshold selection varies depending on the situation.
Therefore, AUROC, which allows for measuring general performance, is commonly employed in tasks like those in this section.

\subsection{Results of the Generative Model Comparison Experiment}\label{sec:res-gen-mod}

\begin{figure}[hbtp]
  \centering
  \includegraphics[keepaspectratio, scale=0.45]{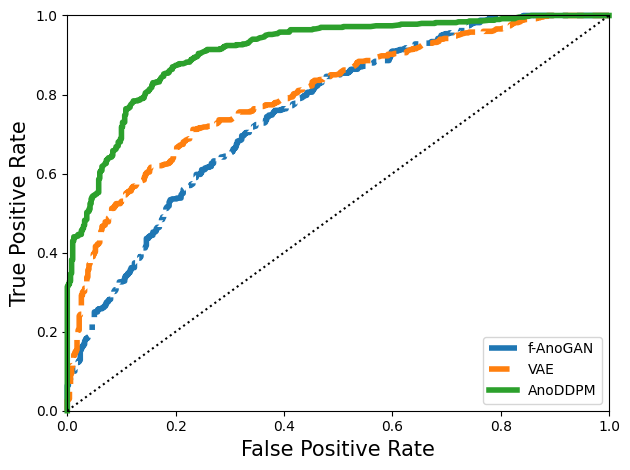}
  \caption{ROC curves for generative models.}
  \label{fig:auroc}
\end{figure}

First, when plotting the ROC curves for VAE, f-AnoGAN, and AnoDDPM trained only on defect-free images, the ROC curves are as shown in Figure \ref{fig:auroc}.

The AUROC values were as follows: VAE achieved $0.800$, f-AnoGAN achieved $0.755$, and AnoDDPM achieved $0.909$.
This indicates that AnoDDPM exhibited the highest performance among the three models.

Table 2 shows the restored and difference images by each model when inputted defective image.
\begin{table*}[tb]
  \centering
   \caption{Restored and difference images of waveforms by each generative model.}
   \label{tab:in-out-diff}
  \begin{tabular*}{15cm}{@{\extracolsep{\fill}}|C|CCC|} \hline
    & Input image & Restored image & Difference image \\ \hline
    VAE & 
    \vspace{0.25cm}\includegraphics[width=0.8\linewidth]{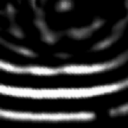} & 
    \vspace{0.25cm}\includegraphics[width=0.8\linewidth]{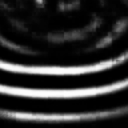} & 
    \vspace{0.25cm}\includegraphics[width=0.8\linewidth]{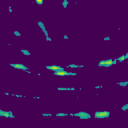} \\ \hline
    f-AnoGAN & 
    \vspace{0.25cm}\includegraphics[width=0.8\linewidth]{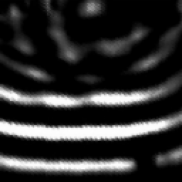} & 
    \vspace{0.25cm}\includegraphics[width=0.8\linewidth]{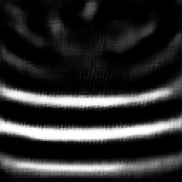} & 
    \vspace{0.25cm}\includegraphics[width=0.8\linewidth]{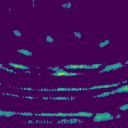} \\ \hline
    AnoDDPM & 
    \vspace{0.25cm}\includegraphics[width=0.8\linewidth]{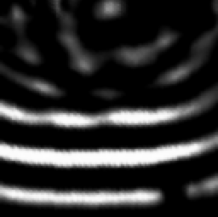} & 
    \vspace{0.25cm}\includegraphics[width=0.8\linewidth]{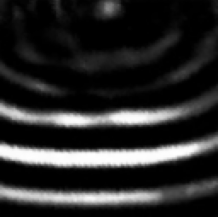} & 
    \vspace{0.25cm}\includegraphics[width=0.8\linewidth]{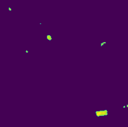} \\ \hline
  \end{tabular*}
\end{table*}
The input images show defects in the bottom right, where the waves exhibit anomalies upon encountering the defect.
Additionally, in the restored images, the wave anomalies appear restored as normal.
And in the difference images, the defect locations are identifiable in all the generative models.
However, in VAE and f-AnoGAN, changes are observed not only in the defective region but also in other areas.
In contrast, AnoDDPM shows specific changes localized to the defective area.
As a result, VAE and f-AnoGAN, which exhibit changes in normal regions, likely yield higher anomaly scores in normal images, leading to the ROC curves depicted in Figure \ref{fig:auroc}.

VAE has been criticized for generating blurry images ~\cite{xu2024maediff,10460109}, and this can be observed in the restored images from VAE, which exhibit noticeable image fuzziness.
Furthermore, GAN has been noted for altering features of normal regions in reconstructed images ~\cite{BAUR2021101952}.
When observing the restored images from f-AnoGAN, it is evident that the shapes of the waves are altered.
It is believed that similar issues occur in this task, where VAE and f-AnoGAN exhibit high anomaly scores due to these problems.

From these, it is evident that AnoDDPM outperforms VAE and f-AnoGAN in terms of defect detection performance.
Therefore, for the comparison with the conventional approach of general object detection algorithms in automated LUVT inspection, AnoDDPM was chosen.

\subsection{Comparison Experiment between Proposed Method and General Object Detection Algorithm}

In the comparison experiment between the proposed method and the general object detection algorithm, the dataset used for training and testing was structured as shown in Table \ref{tab:dataset}.
And after preparing the dataset in five patterns as shown in Table \ref{tab:dataset}, we repeated proposed method five times. 
In this section, we recorded the average of five Precision and five Recall.
\begin{table}[hbtp]
  \caption{Configuration of the dataset. Number of videos in the training and testing datasets.}
  \label{tab:dataset}
  \centering
  \scalebox{0.85}[0.85]{ 
  \begin{tabular}{c|cc}
    \hline
      & Defective & Defect-free \\ \hline
      Training set(Object detection) & 73 & 5 \\ 
      Training set(AnoDDPM) & 78 & 0 \\
      Testing  set & 50 & 50 \\ \hline
  \end{tabular}
  }
\end{table}

\subsubsection{Experimental Settings for the General Object Detection Algorithm}

We used RTMDet~\cite{lyu2022rtmdet}, RetinaNet~\cite{Lin_2017_ICCV}, and YOLOv3~\cite{YOLOv3} as supervised general object detection algorithms.
Also, each model used the parameters shown in Table \ref{tab:parm-od} for training.
\begin{table}[hbtp]
  \caption{Parameters used for training general object detection algorithms.}
  \label{tab:parm-od}
  \centering
  \scalebox{0.8}[0.8]{ 
  \begin{tabular}{c|ccc} \hline
     & RTMDet & RetinaNet  &  YOLOv3 \\ \hline
     Epoch & 100 & 100  & 100 \\
     Batch size & 32 & 8   & 64 \\
     Optimiser & AdamW & SGD  & SGD \\
     Learning rate & 0.004 & 0.01 & 0.001 \\ \hline
  \end{tabular}
  }
\end{table}

\begin{figure}[H]
  \centering
  \includegraphics[keepaspectratio, scale=0.5]{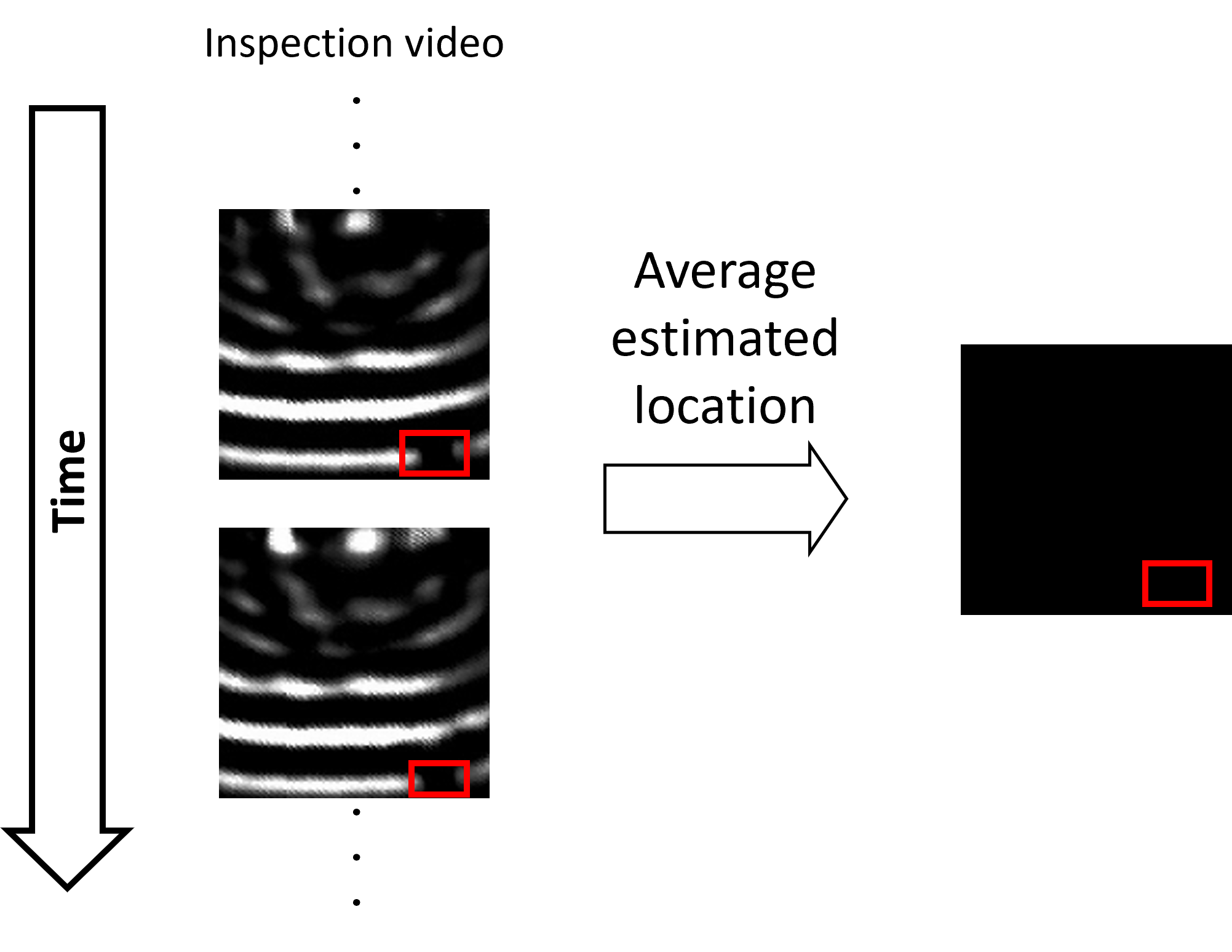}
  \caption{Method for defect localization in a single video using general object detection algorithm.}
  \label{fig:object-detection-video}
\end{figure}

During training, the training data was split into 80\% for training and 20\% for validation.
Then, during each epoch, the loss on the validation data was recorded, and the weights corresponding to the minimum loss were obtained as the learned model weights.
Afterwards, the obtained weights from the training were used to perform defect localization on each time-series image.
We averaged the coordinates estimated from each time-series image.
Then, we evaluated the defect localization as a single video (as shown in Figure \ref{fig:object-detection-video}).

\begin{figure*}[t]
  \begin{tabular}{ll}
    \hspace{0.825cm} (a) & \hspace{0.9cm} (b) \\
    \begin{minipage}[t]{0.45\hsize}
      \centering
      \includegraphics[keepaspectratio, scale=0.5]{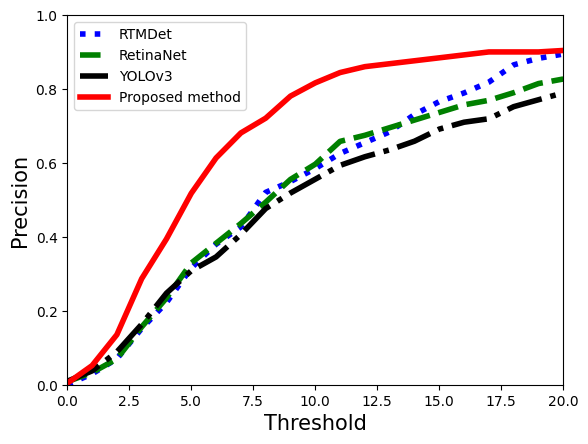}
    \end{minipage} & \vspace{0.5cm}
    \begin{minipage}[t]{0.45\hsize}
      \centering
      \includegraphics[keepaspectratio, scale=0.5]{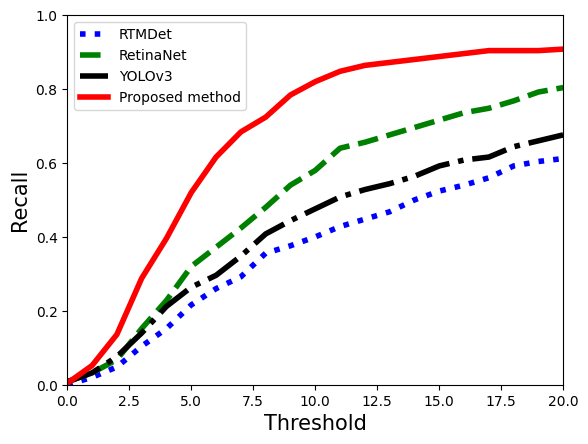}
    \end{minipage}
  \end{tabular}
 \caption{Comparison of defect detection and localization performance between general object detection algorithms and the proposed method. (a) represents Precision, and (b) represents Recall. In both cases, the proposed method shows higher defect detection performance compared to the conventional approach.}
 \label{fig:res-od-proposal}
\end{figure*}

\subsubsection{Experimental Settings for the Proposed Method}

For the proposed method, the generative model AnoDDPM was selected based on the results of Section \ref{sec:res-gen-mod}.
Additionally, the experimental settings were the same as those described in Section \ref{com-gen-models-pre-ddpm}.

\subsection{Evaluation Metrics for the Comparison Experiment between the Proposed Method and General Object Detection Algorithms}

For evaluation, following the approach used in prior research on defect localization in LUVT~\cite{auto-ndt-surp-nakajima}, each model's predictions were classified as true positive (TP) if the predicted defect positions on the testing data were within a distance of $r$ pixels from the actual defect positions.
We used Precision and Recall as the evaluation metrics.

First, Precision is defined using True Positives (TP) and False Positives (FP) as follows:
\begin{align}
\text{Precision} = \frac{\text{TP}}{\text{TP} + \text{FP}}. \label{eq:precision}
\end{align}  
Precision is a metric that indicates the proportion of predicted positives (defects predicted to be present) that are actually true positives (cases where defects are truly present).
This evaluation metric allows us to assess the extent of false detections of defects, with higher values indicating fewer false detections.

On the other hand, Recall is defined using true positives and false negatives (FN) as follows:
\begin{align}
  \text{Recall} = \frac{\text{TP}}{\text{TP} + \text{FN}}. \label{eq:recall}
\end{align}  
Recall is a metric that indicates how well the model can correctly predict the presence of defects among the actual defective data.
This metric allows us to assess the extent of missed defects, with higher values indicating fewer missed detections.

\section{Experimental Results for the General Object Detection Algorithms}

Figure \ref{fig:res-od-proposal} shows the performance comparison between the proposed method with the general object detection algorithms (RTMDet, RetinaNet, and YOLOv3).
In Figure \ref{fig:res-od-proposal}, the horizontal axis represents the threshold of distance, which corresponds to the error between the predicted defect positions and the ground truth defect positions.

In Figure \ref{fig:res-od-proposal}(a), when a threshold of distance is 10 pixels, the proposed method achieved 0.817, RTMDet achieved 0.585, RetinaNet achieved 0.597, and YOLOv3 achieved 0.556 about Precision.
Firstly, among the conventional general object detection algorithms, RetinaNet exhibited the highest performance.
Next, when comparing the proposed method to the conventional object detection algorithm (RetinaNet), the proposed method showed a 22.0\% improvement.
In other words, false detections have been reduced compared to conventional approaches.

Furthermore, in Figure \ref{fig:res-od-proposal}(b), when a threshold of distance is 10, the proposed method achieved 0.820, RTMDet achieved 0.400, RetinaNet achieved 0.580, and YOLOv3 achieved 0.476 about Recall.
Firstly, among the conventional general object detection algorithms, RetinaNet exhibited the highest performance.
Next, when comparing the proposed method with the general object detection algorithm (RetinaNet), the proposed method showed a 24.0\% improvement.
In other words, missed detections have been reduced compared to conventional approaches.

\begin{figure}[htbp]
  \begin{tabular}{cc}
    \multicolumn{2}{c}{ (a) Examples of false detection} \\
    \multicolumn{2}{c}{\hspace{0.85cm} for the conventional approach} \\
    \begin{minipage}[t]{0.45\hsize}
      \centering
      \includegraphics[keepaspectratio, scale=1.25]{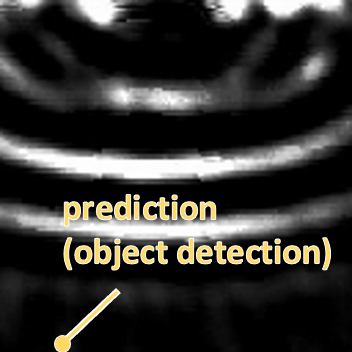}
    \end{minipage} &
    \begin{minipage}[t]{0.45\hsize}
      \centering
      \includegraphics[keepaspectratio, scale=1.25]{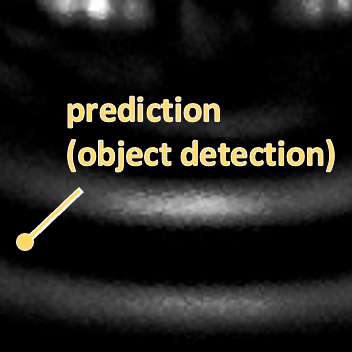}
    \end{minipage} \\
    \multicolumn{2}{c}{ (b) Examples of missed detection} \\
    \multicolumn{2}{c}{\hspace{0.55cm}  for the conventional approach} \\
    \begin{minipage}[t]{0.45\hsize}
      \centering
      \includegraphics[keepaspectratio, scale=1.25]{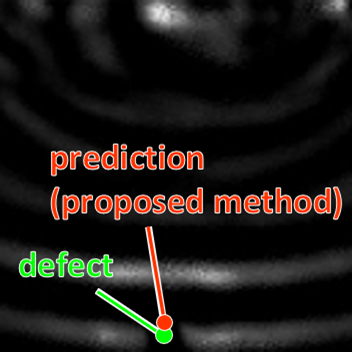}
    \end{minipage} &
    \begin{minipage}[t]{0.45\hsize}
      \centering
      \includegraphics[keepaspectratio, scale=1.25]{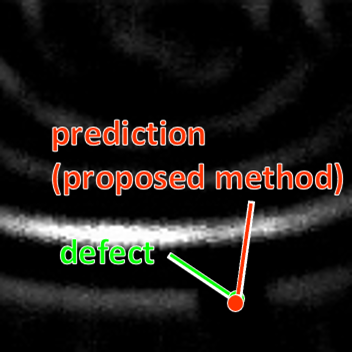}
    \end{minipage} \\
    \multicolumn{2}{c}{ (c) Examples of large position estimation errors} \\
    \multicolumn{2}{c}{\hspace{-1.8cm}  for the conventional approach} \\
    \begin{minipage}[t]{0.45\hsize}
      \centering
      \includegraphics[keepaspectratio, scale=1.25]{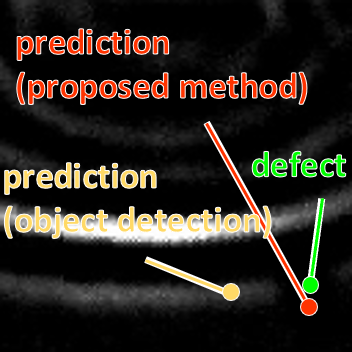}
    \end{minipage} &
    \begin{minipage}[t]{0.45\hsize}
      \centering
      \includegraphics[keepaspectratio, scale=1.25]{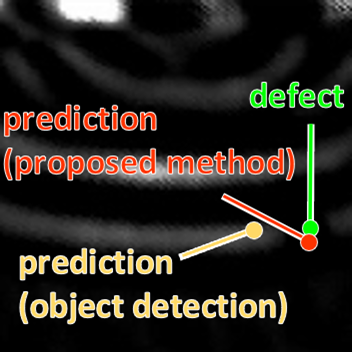}
    \end{minipage} \\
  \end{tabular}
 \caption{Examples of defect detection using the proposed method and the conventional approach. Examples (a) depict cases where the conventional approach resulted in false detections but was correctly judged by the proposed method. Examples (b) depict cases where the conventional approach resulted in missed detections but was correctly identified by the proposed method. Examples (c) depict cases where the conventional approach has a large error in position estimation, but the proposed method is able to estimate the position more accurately.}
 \label{fig:ex-prop-conv}
\end{figure}

Next, let us qualitatively observe the differences between the proposed method and the conventional approach.
Figure \ref{fig:ex-prop-conv} illustrates examples of defect detection using the proposed method and the conventional approach.
Figure \ref{fig:ex-prop-conv}(a) shows the examples of detection for defect-free videos, while Figures \ref{fig:ex-prop-conv}(b) and \ref{fig:ex-prop-conv}(c) depict examples of detection for videos with defects.
Observing Figure \ref{fig:ex-prop-conv}(a), we can see that although there are no defects present, the conventional approach incorrectly detected a defect, whereas the proposed method made the correct judgment.
Observing Figure \ref{fig:ex-prop-conv}(b), we can see that despite the presence of a defect, the conventional approach missed detecting the defect as present, whereas the proposed method correctly identified the defect.
The conventional approach relies on the assumption that a large amount of data with and without defects is necessary for training.
Therefore, when it is difficult to acquire data with defects, as in this task, the performance is not sufficient and the defect detection performance is considered to be low.
Observing Figure \ref{fig:ex-prop-conv}(c), we can see that for videos with defects, the conventional approach exhibits a larger error in defect localization, whereas the proposed method achieves more accurate defect localization.
The lower accuracy in defect localization by the conventional approach is likely due to insufficient training data, which affects the precision of the position estimation.

\section{Conclusion}

This paper proposes an algorithm for defect localization using unsupervised anomaly detection methods, specifically employing generative models, in LUVT.
In LUVT, it is particularly challenging to collect positive examples, which are images containing defects.
As a result, the conventional approach relying on supervised object detection algorithms that require both positive and negative examples suffers from low performance.
Therefore, we focused on anomaly detection methods based on generative models that do not require positive examples in the training data.
Additionally, among generative models, we selected AnoDDPM, which exhibits minimal changes in normal regions and highlights anomalies uniquely.
In this paper, we proposed a model not only for defect detection but also for defect localization to enhance the efficiency of inspection tasks.
We experimentally confirmed that our proposed method improves defect detection and localization compared to general object detection algorithms.
It is worth noting that the materials used in this experiment are limited to aluminum, but we are considering applying the method to composite materials such as concrete or CFRP in the future.

\vspace{0.2cm}
\noindent\textbf{Acknowledgment}
\vspace{0.2cm}
\\

This study was supported by the Sasakawa Science Research Grant from the Japan Science Society.
And supported by a general research grant from the SECOM Science and Technology Promotion Foundation and a Grant-in-Aid for Scientific Research (C) (21K0423100).
In addition, Kyoto University and Hokkaido University, which are the Joint Usage/Collaborative Research Center for Interdisciplinary Large-scale Information Infrastructure, provided computing resources (Project ID: jh220033).

\bibliographystyle{unsrt}
\bibliography{reference}

\end{document}